\documentclass[conference,12pt,onecolumn,draftclsnofoot]{IEEEtran}

\usepackage[utf8]{inputenc}
\usepackage{geometry}
\usepackage{graphicx}
\usepackage{booktabs}
\usepackage{hyperref}
\usepackage{float}
\usepackage{cite}

\hypersetup{
    colorlinks=true,
    linkcolor=blue,
    filecolor=magenta,      
    urlcolor=cyan,
    citecolor=blue,
}

\usepackage{xcolor}
\usepackage{pgfplots}
\pgfplotsset{compat=1.18}

\definecolor{cb-orange}{HTML}{E69F00}
\definecolor{cb-sky}{HTML}{56B4E9}
\definecolor{cb-green}{HTML}{009E73}

\usepackage{tabularx}
\usepackage{microtype}
\usepackage{color}
\usepackage{xspace}
\usepackage{amsmath}
\usepackage{amssymb}
\usepackage{amsthm}
\usepackage{tikz}
\usepackage{url}
\usepackage{enumitem}

\usepackage{titlesec}
\renewcommand{\thesection}{\arabic{section}}
\renewcommand{\thesubsection}{\arabic{section}.\arabic{subsection}}
\renewcommand{\thesubsubsection}{\arabic{section}.\arabic{subsection}.\arabic{subsubsection}}

\titleformat{\section}
  [hang]
  {\normalfont\bfseries\Large}
  {\thesection.}
  {1em}
  {}

\titleformat{\subsection}
  [hang]
  {\normalfont\bfseries\large}
  {\thesubsection.}
  {1em}
  {}

\titleformat{\subsubsection}
  [hang]
  {\normalfont\bfseries}
  {\thesubsubsection.}
  {1em}
  {}

\title{Prognosis Without Prediction: The Intrinsic Geometry of Survival}

\author{
    \IEEEauthorblockN{\large Atahan Karagöz}
    \IEEEauthorblockA{
        \textit{Department of Computer Science} \\
        \textit{University of Basel} \\
        Basel, Switzerland \\
        atahan.karagoez@stud.unibas.ch
    }
}

\begin{document}

\maketitle

\begin{abstract}
\noindent Modeling systemic failure in high-dimensional environments is fundamentally limited by supervised algorithms that attempt to predict sparse, noisy or censored endpoints. To break this reliance on explicit outcome regression, we engineered the Self-Organizing Survival Manifolds (SOSM) framework, a deep learning architecture that decouples unsupervised representation learning from survival labels and treats time-to-event progression as an emergent geometric property. The system projects heterogeneous multi-omics arrays into a unified latent space governed by a pseudotime-weighted, curvature-minimizing graph Laplacian, organizing patient trajectories along low-dissipation geodesic flows. To validate the framework, we applied this architecture to a 184-patient clear cell renal cell carcinoma multi-omics cohort. Across a deterministic 5-fold cross-validation pipeline, SOSM achieved a mean Concordance Index of $0.768 \pm 0.068$, while out-of-fold risk stratification yielded significant survival divergence ($\text{log-rank } p < 0.001$) accompanied by a $14.7\%$ increase in local manifold dissipation among high-risk patients. Furthermore, interrogating the network's encoder weights confirmed that the model autonomously recovered canonical metabolic and microenvironmental drivers of disease progression without supervised feature selection. By demonstrating that systemic failure can be mapped as a geometric phase transition, SOSM establishes a generalizable computational engine for prognostic modeling across biological, industrial and financial complex systems.
\end{abstract}


\section{Introduction}%
\label{sec:introduction}

Mapping the survival and failure trajectories of complex systems remains a fundamental challenge across both academia and industry. Whether anticipating the mortality of biological organisms, the degradation of mechanical networks or the default of financial assets, analysts frequently face the High-Dimension, Low-Sample-Size (HDLSS) trap~\cite{johnstone2009statistical}. In these environments, systems are characterized by thousands of interacting features, yet the number of observed failure events remains inherently small.

Historically, predictive modeling has relied heavily on supervised frameworks, extending from foundational proportional hazards models~\cite{cox1972regression} to modern deep learning architectures~\cite{katzman2018deepsurv, lee2018deephit}. However, these approaches are notoriously susceptible to catastrophic overfitting when exposed to high-dimensional datasets~\cite{herrmann2021large}. Furthermore, supervised models treat survival as an external annotation to be mathematically regressed, ignoring the reality that real-world time-to-event data is frequently noisy, right-censored and subject to systemic collection biases. Conversely, unsupervised representation techniques, such as factor analysis~\cite{argelaguet2018mofa} and network fusion~\cite{wang2014similarity}, successfully address the dimensionality issue but typically discard critical temporal dynamics in favor of bulk variance.

Recent self-supervised contrastive frameworks~\cite{chen2020simple} have attempted to bridge this gap across disparate data modalities~\cite{du2023scccl, ke2021conan}. Similarly, recent multimodal contrastive pre-training and collaborative graph-transformer architectures have demonstrated strong gains in cancer survival analysis~\cite{wang2025pomp, wen2026coformersurv}, yet they still rely on downstream supervised fine-tuning or direct Cox hazard regression during representation learning. Our prior architecture, OmicsCL, successfully demonstrated the utility of contrastive learning for autonomous subtype discovery~\cite{karagoz2025omicsclunsupervisedcontrastivelearning}, yet it fundamentally treated survival as a static grouping rather than a continuous trajectory. While manifold learning and optimal transport have effectively modeled continuous biological transitions at the cellular level~\cite{schiebinger2019optimal, moon2019visualizing}, they are rarely scaled to systemic macroscopic failure. Building on the thermodynamic principles of computational inertia~\cite{karagoz2025computationalinertiaconservedquantity} and minimal entropy generation~\cite{schneider1994life}, we recently conceptualized the Self-Organizing Survival Manifolds (SOSM) theory~\cite{karagoz2025selforganizingsurvivalmanifoldstheory}. SOSM hypothesizes that complex systems evolve along low-curvature geodesic flows within a latent space. Systemic failure corresponds to a geometric phase transition driven by thermodynamic instability.

\begin{figure}[htbp]
    \centering
    \includegraphics[width=0.8\textwidth]{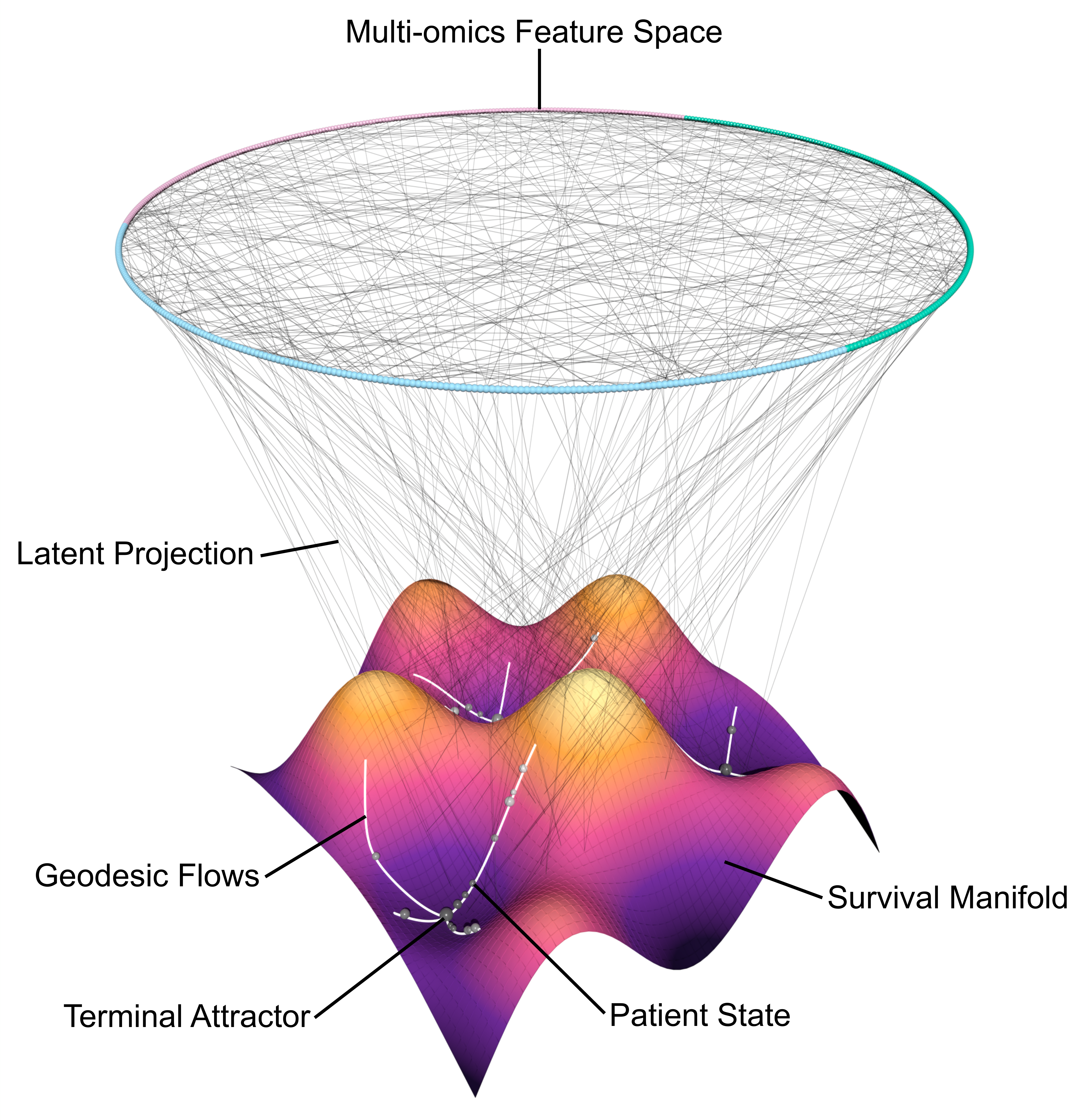}
    \caption{Conceptual topology of the Self-Organizing Survival Manifolds (SOSM) framework. High-dimensional molecular features (top ring) undergo a latent projection via a pseudotime-weighted graph Laplacian. This explicitly forces the neural network to construct a curvature-minimized surface (bottom), where individual patient states naturally descend along stable geodesic flows toward a terminal attractor.}%
    \label{fig:3d_manifold}
\end{figure}

To provide a comprehensive empirical validation of this theory, we selected clinical oncology as our primary testbed (specifically, clear cell Renal Cell Carcinoma, a disease notorious for divergent patient outcomes~\cite{ricketts2018cancer, linehan2013metabolic}). We introduce an unsupervised deep learning pipeline that bypasses supervised regression during representation learning, reserving clinical survival annotations strictly for out-of-fold evaluation and optional one-bit polarity calibration. As illustrated in the conceptual topology of Fig.~\ref{fig:3d_manifold}, the network is mathematically forced to construct a curvature-minimized manifold where geodesic proximity intrinsically reflects prognostic similarity, mapping systemic failure as a geometric phase transition.

\section{Materials and Methods}%
\label{sec:methods}

To empirically evaluate the universal applicability of the Self-Organizing Survival Manifolds (SOSM) theory, we developed a deterministic deep learning framework. The methodology is structured to decouple unsupervised representation learning from supervised outcome regression, ensuring that the discovered geodesic trajectories are intrinsically derived from multi-omics structure rather than overfitted to censored labels.

\subsection{Clinical Cohort and Multi-Omics Processing}
We utilized the multi-omics dataset from The Cancer Genome Atlas Kidney Renal Clear Cell Carcinoma (TCGA-KIRC) cohort~\cite{tcga2013comprehensive, hoadley2018cell}. The raw molecular profiles were stringently filtered to yield a completely matched cohort of 184 patients possessing concurrent gene expression (RNA-seq), DNA methylation, microRNA (miRNA-seq) profiles and curated overall survival data. The cohort exhibited an aggregate clinical event rate of 26.09\%.

\begin{figure}[htbp]
    \centering
    \includegraphics[width=0.8\textwidth]{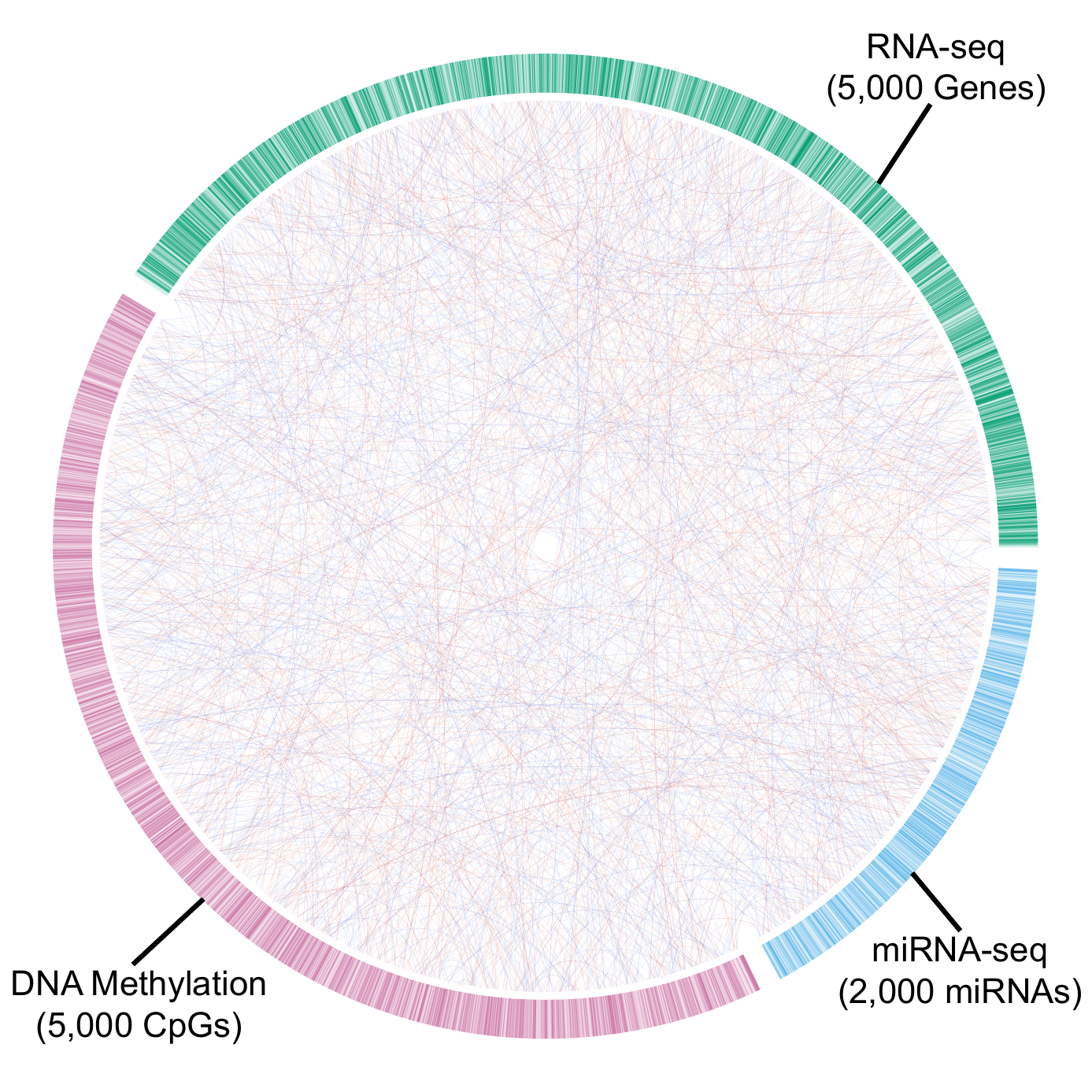}
    \caption{Global connectivity of the high-dimensional multi-omics feature space. This dense bipartite network encompasses 12,000 highly variable features spanning transcriptomic (RNA-seq), epigenomic (DNA Methylation) and post-transcriptional (miRNA-seq) modalities, visually representing the computational complexity of the HDLSS challenge prior to latent manifold projection.}%
    \label{fig:dense_circos}
\end{figure}

To mitigate the curse of dimensionality without introducing data leakage, variance-based feature selection and $Z$-score standardization were fitted strictly on each training split and applied blindly to held-out test folds. Within each fold, the top highly variable features were retained across the modalities: 5,000 for RNA, 5,000 for DNA methylation and 2,000 for miRNA\@. The scale and dense connectivity of this 12,000-feature network are visualized in Fig.~\ref{fig:dense_circos}, illustrating the high-dimensional baseline prior to latent manifold projection. Standardization to zero mean and unit variance prevents higher-magnitude expression platforms from dominating the latent embedding gradients during contrastive backpropagation~\cite{argelaguet2018mofa}.

\subsection{Prognostic Pseudotime Initialization}
In a purely unsupervised survival framework, utilizing exact time-to-death annotations to directly optimize the embedding space violates the premise of label-free discovery. However, biological systems exhibit known transcriptional signatures of progression. Drawing upon the concept of computational inertia and thermodynamic stability~\cite{karagoz2025computationalinertiaconservedquantity}, we generated an initial ``prognostic pseudotime'' proxy ($t$) in a fully unsupervised manner to orient the pseudotime-weighted graph Laplacian.

This proxy was derived via a single-sample Gene Set Enrichment Analysis (ssGSEA) scoring scheme, aggregating canonical proliferation and immune infiltration markers from the RNA modality as $+1.5 \times \text{proliferation} - 1.2 \times \text{immune}$. When ground-truth survival data are available on a training fold, an optional one-bit polarity calibration aligns the global sign of this continuous vector, which is then applied blindly to unseen test cohorts without further label access. Rather than serving as a direct survival surrogate, this raw biological-prior vector provides an uncalibrated heuristic for initial manifold orientation.

\subsection{Neural Architecture and Geometric Projection}
The SOSM architecture employs a late-fusion multi-branch neural network. We defined three independent omics-specific encoders, $f_{rna}$, $f_{methyl}$ and $f_{mirna}$, each parameterized as a multi-layer perceptron (MLP). To enforce a stable geometric projection, each encoder sequence comprises dense linear layers transitioning through intermediate dimensions of 512 and 128, followed by batch normalization and Rectified Linear Unit (ReLU) activations to mitigate internal covariate shift.

The final projection layer maps the modality-specific features into a compact 16-dimensional latent space ($d=16$). To establish a well-defined topological manifold, the output of each encoder is explicitly $L_2$-normalized to lie on the unit hypersphere:
\begin{equation}
z^{(v)} = \frac{f_{\theta}^{(v)}(x)}{\|f_{\theta}^{(v)}(x)\|_2}, \quad v \in \{\text{RNA, Methyl, miRNA}\}.
\end{equation}
The unified patient embedding $Z$ is subsequently derived via an unweighted average of the $L_2$-normalized modality vectors, ensuring that the final survival manifold requires consensus across the disparate biological layers.

\subsection{The SOSM Objective: Alignment and Curvature Regularization}
The network is optimized using a dual-component objective function that balances modality alignment with manifold smoothness. The first component is the Normalized Temperature-scaled Cross-Entropy (NT-Xent) loss~\cite{chen2020simple}, deployed to maximize representational agreement between the independent omics views of the same patient:
\begin{equation}
\mathcal{L}_{align} = \mathcal{L}_{NT-Xent}(z_{rna}, z_{methyl}, \tau) + \mathcal{L}_{NT-Xent}(z_{rna}, z_{mirna}, \tau),
\end{equation}
where the temperature scaling parameter $\tau$ was set to 0.1 to penalize geometric dispersion.

The second component is the Survival Curvature Loss ($\mathcal{L}_{curv}$). Guided by the principle that biological stability maps to low-curvature geodesic flows~\cite{karagoz2025selforganizingsurvivalmanifoldstheory}, we constructed a Gaussian kernel $W$ based on the pseudotime proxy $t$:
\begin{equation}
W_{ij} = \exp\left(-\frac{|t_i - t_j|^2}{2\sigma^2}\right),
\end{equation}
with bandwidth $\sigma = 1.0$. We then computed the normalized graph Laplacian $L = I - D^{-1/2} W D^{-1/2}$, where $D$ is the diagonal degree matrix. The curvature penalty is defined as the squared Frobenius norm of the Laplacian projection:
\begin{equation}
\mathcal{L}_{curv} = \| L Z \|_F^2.
\end{equation}
This term geometrically forces patients exhibiting similar pseudotime signatures to align closely within the 16-dimensional latent space, minimizing local curvature and establishing a smooth trajectory gradient~\cite{karagoz2025selforganizingsurvivalmanifoldstheory}. By anchoring the high-dimensional features to this stable topological surface (Fig.~\ref{fig:3d_manifold}), the network ensures that individual patient trajectories follow low-dissipation paths rather than scattering chaotically. The total training objective combines alignment and curvature regularization:
\begin{equation}
\mathcal{L}_{total} = \mathcal{L}_{align} + \lambda_{curv} \mathcal{L}_{curv},
\end{equation}
where $\lambda_{curv} = 0.1$ acts as a rigidity coefficient governing the manifold constraints.

\subsection{Evaluation and Statistical Inference}
To prevent data leakage and evaluate out-of-sample generalizability, the entire SOSM framework was executed within a deterministic 5-fold cross-validation pipeline (global seed fixed at 3 across CPU and CUDA backends for exact experimental reproducibility). Within each fold, feature selection, standardization, pseudotime polarity calibration and neural network optimization were performed solely on the training split (80\% of the cohort) over 500 epochs.

Once training converged, frozen encoders projected the held-out test split (20\%) onto the 16-dimensional manifold. To evaluate prognostic discrimination, an $L_2$-penalized Cox Proportional Hazards model ($\text{penalizer} = 1.0$) was fitted on the continuous training embeddings and evaluated blindly on the test embeddings using Harrell's Concordance Index (C-index), alongside a purely unsupervised K-Means clustering baseline ($k=2$, $n_{init}=10$). Patients were stratified into Low-Risk and High-Risk cohorts at the 75th percentile of predicted partial hazard scores, and survival separation was tested via the two-sample Mantel-Cox log-rank test. To quantitatively test the thermodynamic phase-transition hypothesis on unseen patients, local manifold curvature and entropy dissipation were quantified per patient as the Euclidean norm of the test-set graph Laplacian projection, $\delta_i = \| (L_{\text{test}} Z_{\text{test}})_i \|_2$, comparing Low-Risk homeostatic states against High-Risk collapsed states. Finally, biological interpretability was established by ranking features by the mean absolute weights of each primary encoder layer and computing cosine similarity across linearized input-to-latent sensitivity vectors ($W_{\text{proj}} W_{h2} W_{h1}$) to identify the top 5\% cross-modality co-activation pairs.

\section{Results}%
\label{sec:results}

\subsection{Prognostic Robustness and Geometric Convergence}
\begin{figure}[htbp]
    \centering
    \includegraphics[width=0.8\textwidth]{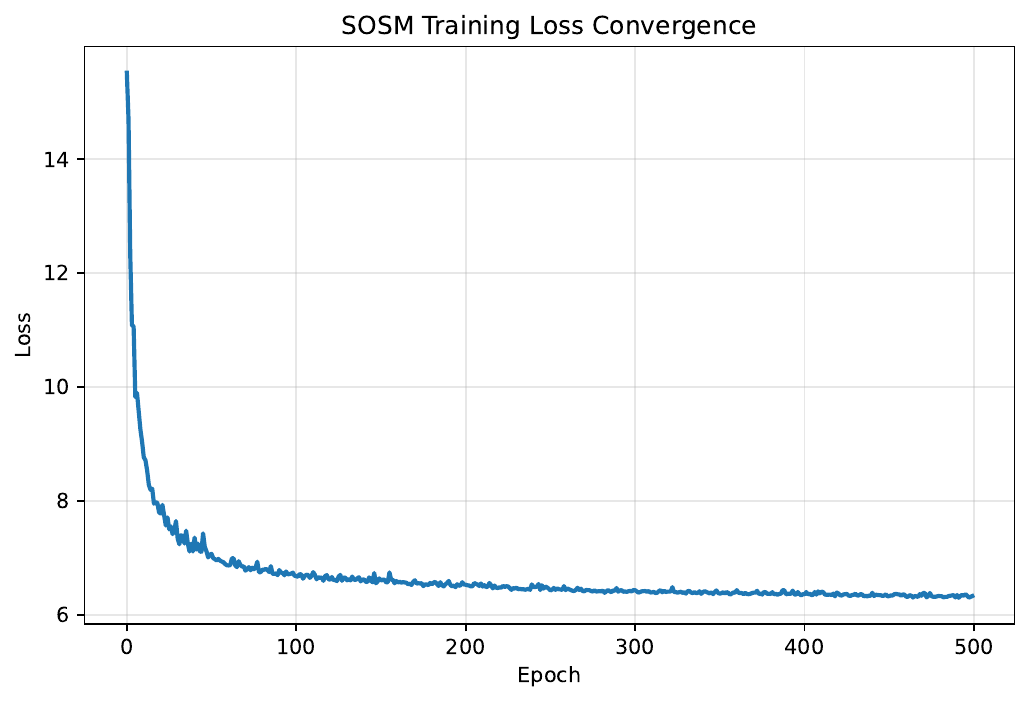}
    \caption{SOSM training loss trajectory for the optimal cross-validation fold.}%
    \label{fig:training_loss}
\end{figure}

By decoupling representation learning from explicit survival regression, the SOSM framework circumvented the severe overfitting characteristic of HDLSS environments. Throughout multi-fold optimization, the neural network exhibited stable convergence dynamics. Optimizing the dual-component objective to balance NT-Xent modality alignment with the curvature penalty produced a smooth, monotonic decrease in training loss over 500 epochs without signs of latent divergence or representational collapse (Fig.~\ref{fig:training_loss}).

\begin{figure}[htbp]
    \centering
    \includegraphics[width=0.8\textwidth]{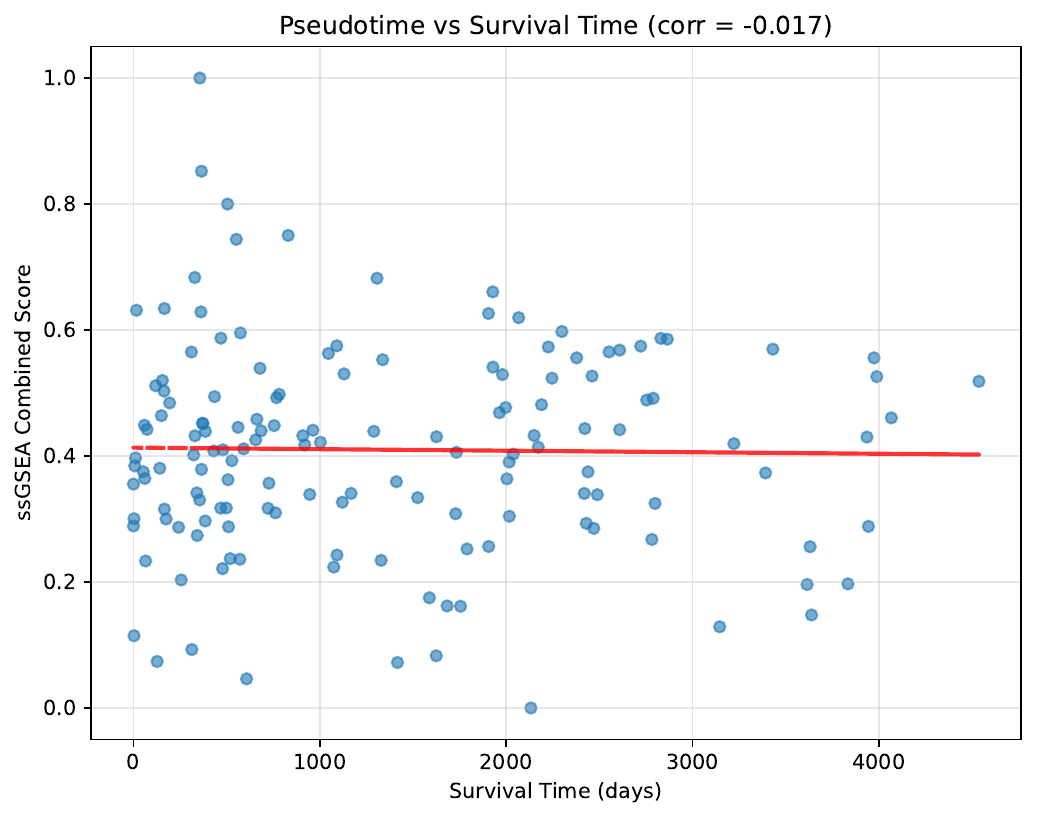}
    \caption{Correlation diagnostic between the raw ssGSEA prognostic pseudotime proxy and clinical survival time.}%
    \label{fig:correlation_diagnostic}
\end{figure}

To ensure the network was guided by intrinsic biological priors rather than explicit event annotations, manifold initialization relied on an uncalibrated prognostic pseudotime proxy derived from canonical proliferation and immune transcriptomic markers. Diagnostic evaluation of this ssGSEA-based proxy confirmed near-zero linear correlation with raw clinical survival time ($r = -0.017$, Fig.~\ref{fig:correlation_diagnostic}), verifying that the initial kernel heuristic does not act as a trivial surrogate for outcome labels.

From this minimal biological anchor, the SOSM architecture generalized robustly to unseen patient cohorts. Evaluating the continuous 16-dimensional latent representations via a penalized Cox Proportional Hazards model across all held-out test folds yielded a mean Concordance Index (C-index) of $0.768 \pm 0.068$ (reaching $0.848$ on the optimal fold), substantially outperforming naive discrete K-Means partitioning (mean C-index $= 0.507$). This cross-validated discrimination confirms that the low-curvature geodesic flows learned by the encoders capture continuous prognostic structure rather than stochastic memorization.

\subsection{Continuous Gradients and Phase Separation of the Survival Manifold}
While the mean C-index validates prognostic accuracy, the core principle of SOSM is that clinical progression emerges from macroscopic topological organization in the biological state space. To interrogate this geometry, we projected the 16-dimensional latent embeddings of the optimal cross-validation fold onto two dimensions using Uniform Manifold Approximation and Projection (UMAP).

\begin{figure}[htbp]
    \centering
    \includegraphics[width=0.8\textwidth]{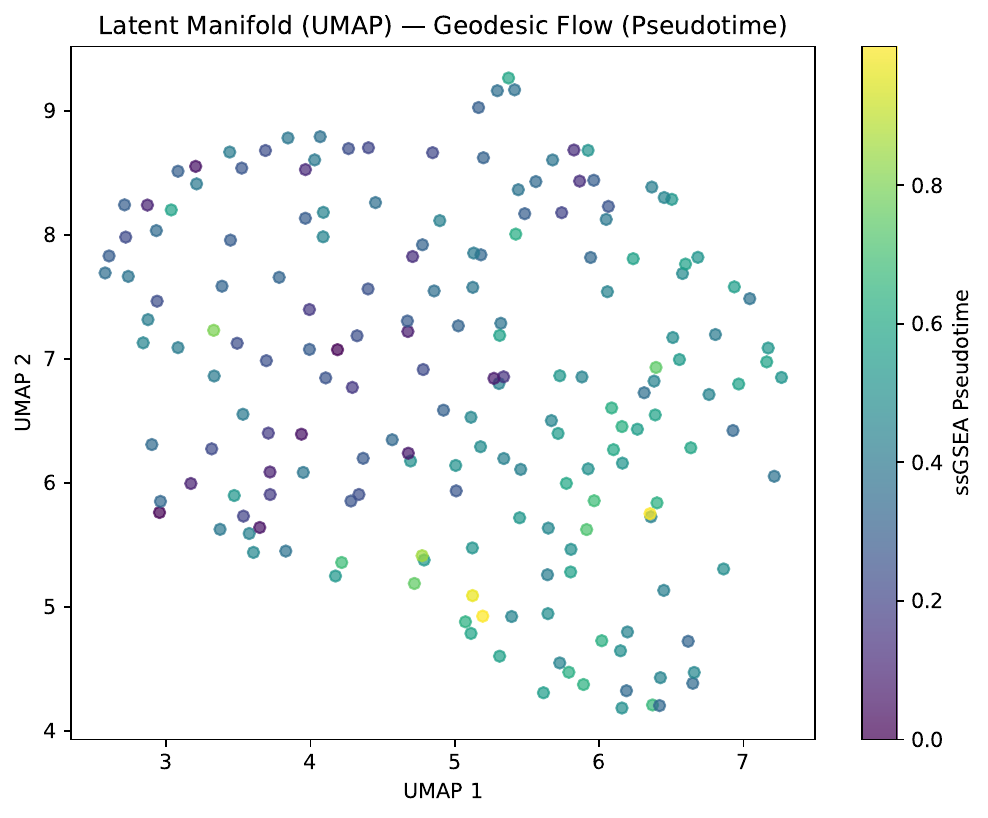}
    \caption{UMAP projection of the learned latent space colored by the continuous ssGSEA pseudotime proxy. The smooth spatial progression illustrates how curvature regularization organizes multi-omics patient embeddings along a continuous manifold while preserving biological heterogeneity.}%
    \label{fig:umap_ssgsea}
\end{figure}

The resulting projections reveal the dual structure of the survival manifold. First, when colored by the continuous ssGSEA pseudotime proxy (Fig.~\ref{fig:umap_ssgsea}), the manifold exhibits a coherent, progressive gradient across the patient cohort. Minimizing Laplacian curvature energy forces the latent space to organize along a continuous biological trajectory, smoothly transitioning across prognostic states without local fragmentation.

\begin{figure}[htbp]
    \centering
    \includegraphics[width=0.8\textwidth]{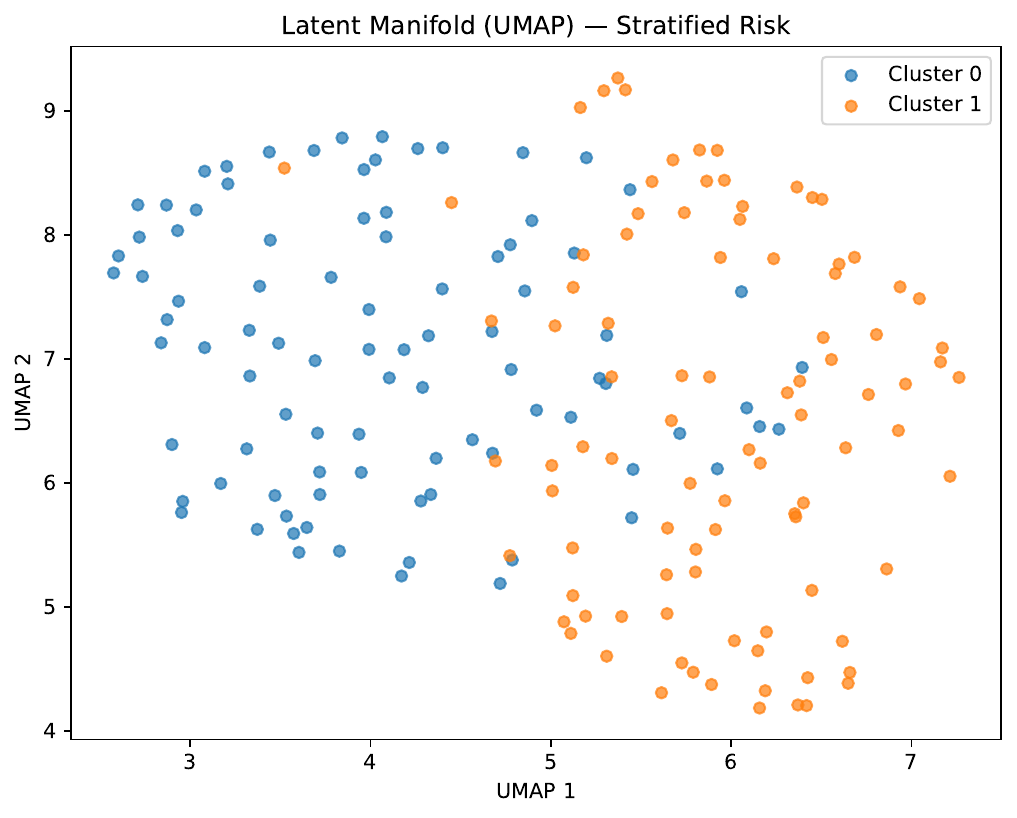}
    \caption{UMAP projection of the learned latent space colored by unsupervised K-Means cluster assignments. The partition illustrates the geometric phase separation between homeostatic and high-progression patient regimes.}%
    \label{fig:umap_clusters}
\end{figure}

Second, unsupervised K-Means clustering ($k=2$) applied directly to the latent embeddings partitions this continuous manifold into two macroscopically distinct regions (Fig.~\ref{fig:umap_clusters}). This structural bifurcation reflects the transition from a homeostatic, low-dissipation state to a high-curvature regime associated with accelerated disease progression.

\begin{figure}[htbp]
    \centering
    \includegraphics[width=0.8\textwidth]{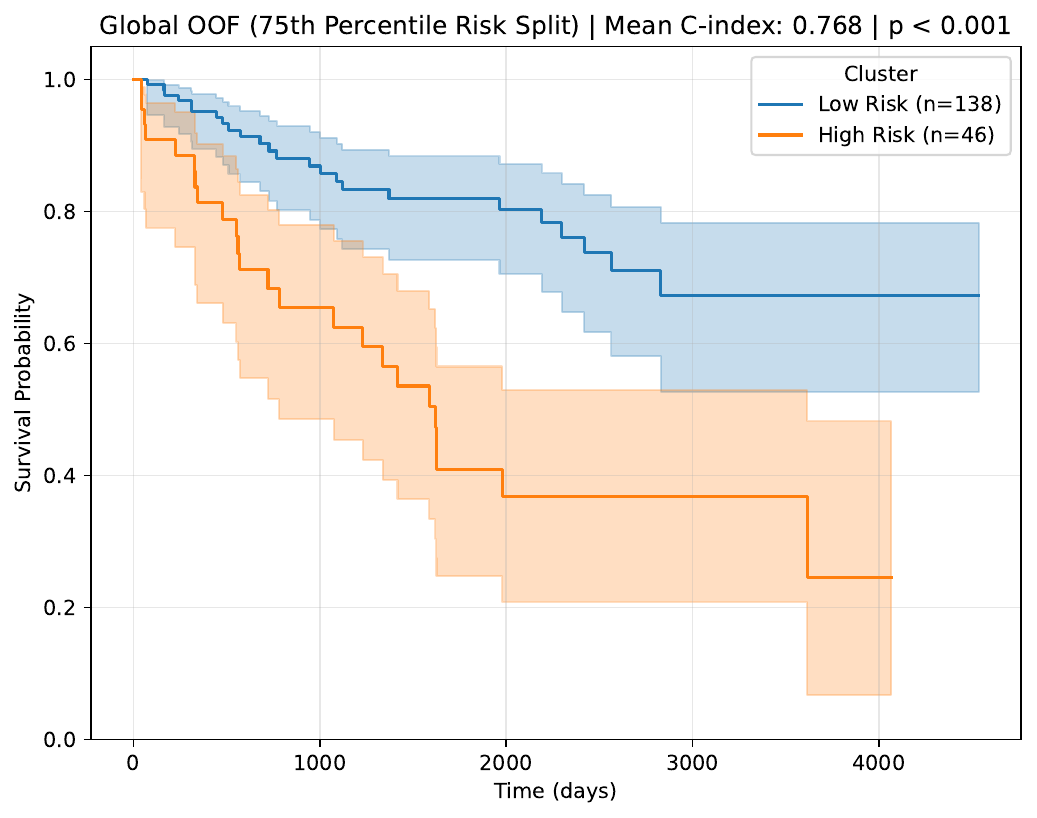}
    \caption{Global out-of-fold Kaplan-Meier survival curves across the full TCGA-KIRC cohort. Shaded bands denote confidence intervals for the stratified Low-Risk and High-Risk trajectories.}%
    \label{fig:km_global_oof}
\end{figure}

\begin{figure}[htbp]
    \centering
    \includegraphics[width=0.8\textwidth]{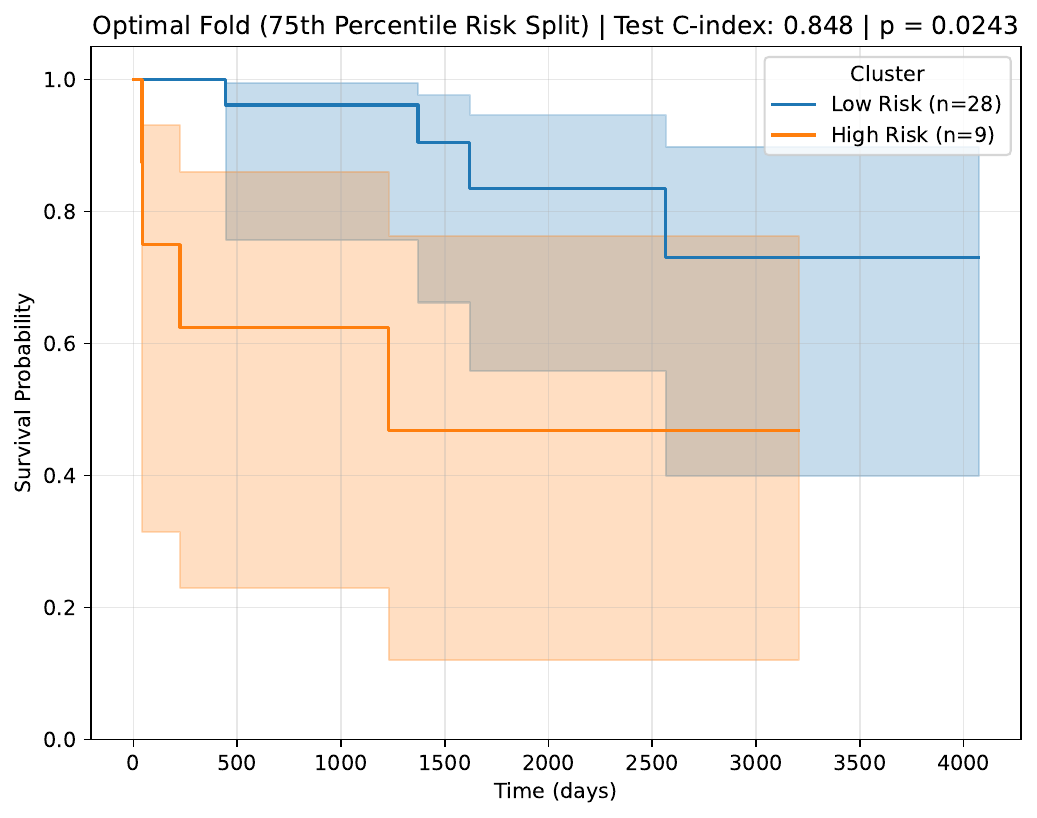}
    \caption{Held-out test set Kaplan-Meier survival curves for the optimal cross-validation fold. Shaded bands denote confidence intervals for the stratified Low-Risk and High-Risk trajectories.}%
    \label{fig:km_test_optimal}
\end{figure}

To evaluate clinical stratification, we examined Kaplan-Meier survival trajectories using the 75th percentile of out-of-fold Cox partial hazard scores across both the concatenated full cohort ($N = 184$, Fig.~\ref{fig:km_global_oof}) and the isolated test split of the optimal fold ($n = 37$, Fig.~\ref{fig:km_test_optimal}). Across the concatenated full cohort, High-Risk patients ($n = 46$) experienced markedly worse overall survival than Low-Risk patients ($n = 138$), achieving significant separation under the Mantel-Cox log-rank test ($p < 0.001$, Fig.~\ref{fig:km_global_oof}). Consistent stratification held within the strictly held-out optimal test fold ($n = 9$ High-Risk vs. $n = 28$ Low-Risk, log-rank $p = 0.0243$, Fig.~\ref{fig:km_test_optimal}), confirming that the geometric boundaries learned by SOSM generalize to unseen patients.

Furthermore, evaluating local Laplacian dissipation across these test embeddings provides direct quantitative support for the thermodynamic phase-transition hypothesis of SOSM\@. Patients in the Low-Risk homeostatic regime exhibited a mean local curvature dissipation of $0.5131$, whereas patients in the High-Risk regime demonstrated an elevated mean dissipation of $0.5887$. This $14.7\%$ increase in local geometric dispersion confirms that clinical mortality risk corresponds to accelerated local entropy production ($\Delta S$) and structural collapse along the survival manifold.

\subsection{Biological Interpretability and Unsupervised Biomarker Discovery}
\begin{figure}[htbp]
    \centering
    \includegraphics[width=\textwidth]{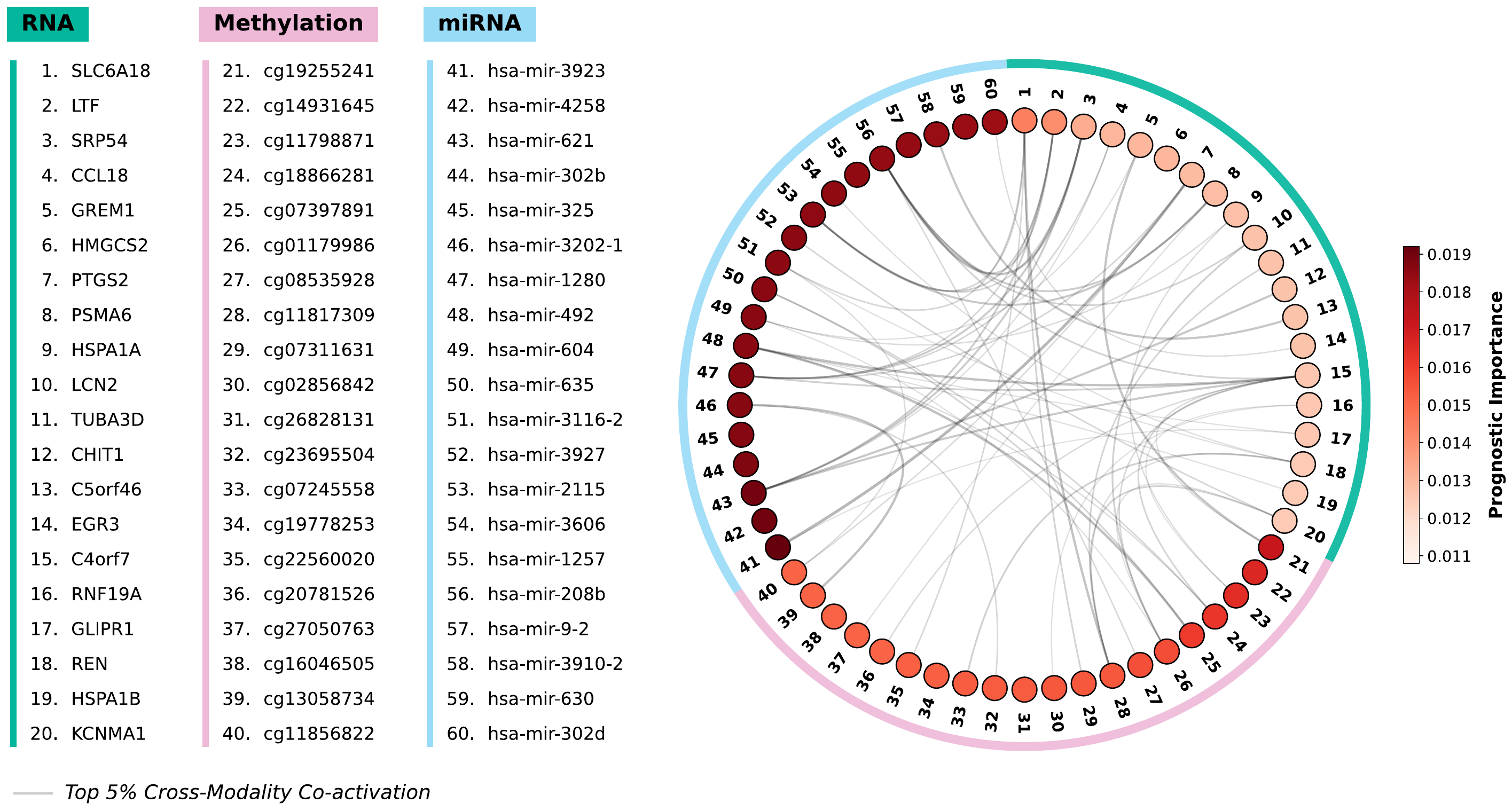}
    \caption{Top 5\% cross-modality co-activations derived from linearized input-to-latent sensitivity across the SOSM encoders. Outer nodes represent the top 20 biomarkers per modality colored by primary encoder attribution weight, while inner chords depict inter-modality alignments within the shared survival manifold.}%
    \label{fig:top5_circos}
\end{figure}

To translate the learned Riemannian survival manifold into actionable molecular mechanisms, we extracted the highest-weighted features from the first linear layer of each modality-specific encoder alongside their strongest cross-modality interactions (Table~\ref{tab:biomarkers} and Table~\ref{tab:coactivation}). Within individual omics layers, weight attribution highlights key regulators of renal solute transport, innate immunity and post-transcriptional control, led by \textit{SLC6A18}, \textit{LTF} and \textit{SRP54} in transcriptomics, \texttt{cg19255241} and \texttt{cg14931645} in DNA methylation and \texttt{hsa-mir-3923} and \texttt{hsa-mir-4258} in miRNA expression (Table~\ref{tab:biomarkers}). Beyond single-omic importance, evaluating linearized input-to-latent sensitivity across the 16-dimensional shared embedding reveals tightly coupled cross-modality axes (Fig.~\ref{fig:top5_circos} and Table~\ref{tab:coactivation}). Notably, \texttt{hsa-mir-621} and \texttt{hsa-mir-492} emerge as prominent convergence hubs, aligning simultaneously with multiple top-ranked transcripts (\textit{C4orf7}, \textit{CHIT1}, \textit{SRP54}) and epigenetic loci (\texttt{cg07397891}) onto shared manifold coordinates.

\begin{table}[htbp]
\centering
\caption{Top 10 Multi-Omics Biomarkers Discovered by the SOSM Architecture. Features are ranked by primary encoder attribution weight, reflecting their relative contribution to the survival manifold.}%
\label{tab:biomarkers}
\resizebox{\textwidth}{!}{%
\begin{tabular}{lrlrlr}
\toprule
\textbf{RNA Gene} & \textbf{RNA Weight} & \textbf{Methyl CpG} & \textbf{Methyl Weight} & \textbf{miRNA Name} & \textbf{miRNA Weight} \\
\midrule
SLC6A18|348932 & 0.0132 & cg19255241 & 0.0167 & hsa-mir-3923   & 0.0192 \\
LTF|4057       & 0.0128 & cg14931645 & 0.0160 & hsa-mir-4258   & 0.0190 \\
SRP54|6729     & 0.0118 & cg11798871 & 0.0158 & hsa-mir-621    & 0.0189 \\
CCL18|6362     & 0.0115 & cg18866281 & 0.0154 & hsa-mir-302b   & 0.0187 \\
GREM1|26585    & 0.0115 & cg07397891 & 0.0153 & hsa-mir-325    & 0.0186 \\
HMGCS2|3158    & 0.0115 & cg01179986 & 0.0148 & hsa-mir-3202-1 & 0.0185 \\
PTGS2|5743     & 0.0113 & cg08535928 & 0.0147 & hsa-mir-1280   & 0.0185 \\
PSMA6|5687     & 0.0112 & cg11817309 & 0.0145 & hsa-mir-492    & 0.0184 \\
HSPA1A|3303    & 0.0111 & cg07311631 & 0.0145 & hsa-mir-604    & 0.0184 \\
LCN2|3934      & 0.0111 & cg02856842 & 0.0145 & hsa-mir-635    & 0.0184 \\
\bottomrule
\end{tabular}%
}
\end{table}

\begin{table}[htbp]
\centering
\caption{Top 10 Cross-Modality Co-Activation Pairs in the Shared Survival Manifold. Pairs are ranked by absolute cosine similarity ($|A_{ij}| \in [0,1]$) of linearized input-to-latent sensitivity across distinct omics modalities (top 5\% threshold $= 0.4790$, $n = 60$ pairs total).}%
\label{tab:coactivation}
\resizebox{\textwidth}{!}{%
\begin{tabular}{lllr}
\toprule
\textbf{Modality Pair} & \textbf{Source Feature} & \textbf{Target Feature} & \textbf{Latent Co-Activation} \\
\midrule
RNA -- miRNA         & PTGS2|5743     & hsa-mir-3923   & 0.7460 \\
Methylation -- miRNA & cg07397891     & hsa-mir-492    & 0.6841 \\
Methylation -- miRNA & cg13058734     & hsa-mir-3202-1 & 0.6765 \\
RNA -- miRNA         & C4orf7|260436  & hsa-mir-492    & 0.6693 \\
RNA -- Methylation   & SLC6A18|348932 & cg11817309     & 0.6625 \\
RNA -- Methylation   & GREM1|26585    & cg19255241     & 0.6538 \\
RNA -- miRNA         & SRP54|6729     & hsa-mir-208b   & 0.6510 \\
RNA -- miRNA         & C5orf46|389336 & hsa-mir-3910-2 & 0.6434 \\
RNA -- miRNA         & CHIT1|1118     & hsa-mir-621    & 0.6390 \\
RNA -- miRNA         & C4orf7|260436  & hsa-mir-621    & 0.6270 \\
\bottomrule
\end{tabular}%
}
\end{table}

The biological identities of these prioritized markers align closely with established hallmarks of clear cell Renal Cell Carcinoma (ccRCC) pathogenesis. Among the top transcriptomic features, \textit{HMGCS2} (3-hydroxy-3-methylglutaryl-CoA synthase 2) encodes a rate-limiting enzyme in ketogenesis whose epigenetic inactivation and downregulation drive metabolic reprogramming and tumor cell motility in aggressive ccRCC~\cite{linehan2013metabolic, wettersten2015grade, han2021epigenetic}. Similarly, the prioritization of inflammatory and stromal remodeling mediators, including \textit{PTGS2} (COX-2)~\cite{tuna2004significance, lee2012cyclooxygenase}, \textit{GREM1}, \textit{LCN2} and the macrophage-associated chemokine \textit{CCL18}, indicates that the manifold captures both tumor-intrinsic metabolic shifts and tumor microenvironment (TME) immune dynamics~\cite{turner2014cytokines}. That the model isolated these coordinated metabolic and microenvironmental regulators through curvature-regularized contrastive alignment without supervised feature selection demonstrates that SOSM decodes fundamental biophysical drivers of tumor progression.

\section{Discussion}%
\label{sec:discussion}

Our findings motivate a reassessment of how complex system trajectories are modeled computationally. While we utilized clinical oncology as a stringent high-dimensional testbed, our results provide empirical support for the Self-Organizing Survival Manifolds (SOSM) theory~\cite{karagoz2025selforganizingsurvivalmanifoldstheory}. By decoupling manifold representation learning from supervised survival regression, we demonstrate that prognostic trajectories can be recovered directly from the intrinsic topology of multi-omics data without overfitting to censored clinical annotations.

The continuous organization of the latent space indicates that complex systems preferentially evolve along low-curvature, energy-efficient pathways, echoing thermodynamic principles of minimal entropy generation~\cite{schneider1994life, karagoz2025energenticintelligenceselfsustainingsystems}. In our cohort, continuous geodesic modeling substantially outperformed naive discrete clustering, while high-risk patients exhibited elevated local Laplacian dissipation. This divergence mirrors the theoretical predictions of computational inertia, where homeostatic structural invariants eventually succumb to chaotic dispersion~\cite{karagoz2025computationalinertiaconservedquantity}, providing quantitative evidence that systemic failure is not a linear decline but a geometric phase transition triggered by accumulating biological stress.

Moving beyond earlier contrastive representation efforts that isolated static patient clusters~\cite{karagoz2025omicsclunsupervisedcontrastivelearning, du2023scccl}, our framework demonstrates that prognostic states reside on a unified topological continuum. A key validation of the SOSM objective is its label-free recovery of domain-specific mechanisms. In our clinical application, curvature-regularized alignment mapped the tumor's biophysical constraints, autonomously isolating canonical regulators of ketogenic metabolic disruption and inflammatory microenvironmental remodeling~\cite{wettersten2015grade, han2021epigenetic, lee2012cyclooxygenase, turner2014cytokines}. Recovering these coordinated multi-omics axes through geometric regularization highlights the potential of SOSM for broader scientific and industrial domains characterized by high-dimensional, noisy and right-censored observations.

Despite these advancements, several limitations warrant consideration. First, the current instantiation relies on a transcriptomic ssGSEA proxy to provide initial biological orientation for the pseudotime-weighted graph Laplacian, alongside optional one-bit polarity calibration on training splits. Replacing this domain-specific heuristic with intrinsic entropic mappings derived directly from optimal transport theory~\cite{villani2008optimal} remains a central direction for future work. Second, while cross-validation on the 184-patient TCGA-KIRC cohort demonstrates strong out-of-sample generalizability, validating these geometric phase transitions across larger multi-center clinical cohorts as well as non-biological industrial and financial systems will be essential to establish universal utility.

\section{Conclusion}%
\label{sec:conclusion}

The realization of the SOSM framework marks a departure from traditional methodologies in survival analysis and predictive modeling. For decades, academia and industry have been constrained by supervised paradigms that coerce high-dimensional data arrays into predicting external, often censored, temporal annotations. By showing that heterogeneous multi-omics data organizes along low-curvature geodesic trajectories without supervised survival regression, we demonstrate that systemic resilience and failure are inherently encoded in the geometric structure of the data itself.

This framework establishes a unified mathematical language that bridges representation learning, biophysics and complex systems theory, laying the groundwork for a generalizable geometric approach to survival modeling. As these architectures scale beyond precision oncology, the modeling of survival can increasingly transition from the statistical regression of arbitrary endpoints to the geometric mapping of complex system dynamics.

\section*{Data and Code Availability}%
The complete SOSM source code, deterministic cross-validation pipeline, and pre-computed evaluation logs are publicly available at \url{https://github.com/Atahanka/SOSM}. Public TCGA-KIRC multi-omics and clinical data are retrieved programmatically from the Broad Institute GDAC Firehose repository (\url{https://gdac.broadinstitute.org/}).

\bibliographystyle{IEEEtran}
\bibliography{refs}

\end{document}